\documentclass[sigconf,nonacm]{acmart}
\acmConference[ASE 2024]{IEEE/ACM International Conference on Automated Software Engineering}{27 October - 1 November, 2024}{California, United States}
\usepackage[utf8]{inputenc} 
\usepackage{url}            
\usepackage{booktabs}       
\usepackage{amsfonts}       
\usepackage{nicefrac}       
\usepackage{microtype}      
\usepackage{xcolor}         

\usepackage{microtype}
\usepackage{graphicx}
\usepackage{tabularx}
\usepackage{booktabs} 
\usepackage[utf8]{inputenc} 
\usepackage{url}
\usepackage{booktabs}    
\usepackage{amsfonts}     
\usepackage{nicefrac}       
\usepackage{microtype}     
\usepackage{utfsym}
\usepackage{float}
\usepackage{color}
\usepackage[most]{tcolorbox}
\usepackage[ruled,vlined,linesnumbered]{algorithm2e}
\usepackage{amsmath}
\usepackage{multirow}
\usepackage{multicol}
\usepackage{algpseudocode}
\usepackage{amssymb}
\usepackage{longtable}
\usepackage{soul}
\usepackage{subcaption}
\usepackage{natbib}
\usepackage{bbding}
\usepackage{indentfirst}
\usepackage{cleveref}

\setcitestyle{numbers,square,comma}


\begin{document}


\title[Automata Extraction from Transformers]{Automata Extraction from Transformers}

\author{Yihao Zhang}
\affiliation{%
    \institution{Peking University}
    \country{China}
}
\email{zhangyihao@stu.pku.edu.cn}

\author{Zeming Wei}
\affiliation{%
    \institution{Peking University}
    \country{China}
}
\email{weizeming@stu.pku.edu.cn}

\author{Meng Sun}
\authornote{Meng Sun is the Corresponding author.}
\affiliation{%
    \institution{Peking University}
    \country{China}
}
\email{sunm@pku.edu.cn}

\begin{abstract}
    In modern machine (ML) learning systems, Transformer-based architectures have achieved milestone success across a broad spectrum of tasks, yet understanding their operational mechanisms remains an open problem. To improve the transparency of ML systems, automata extraction methods, which interpret stateful ML models as automata typically through formal languages, have proven effective for explaining the mechanism of recurrent neural networks (RNNs). However, few works have been applied to this paradigm to Transformer models. In particular, understanding their processing of formal languages and identifying their limitations in this area remains unexplored. In this paper, we propose an automata extraction algorithm specifically designed for Transformer models. Treating the Transformer model as a black-box system, we track the model through the transformation process of their internal latent representations during their operations, and then use classical pedagogical approaches like \textit{L}$^*$ algorithm to interpret them as deterministic finite-state automata (DFA). Overall, our study reveals how the Transformer model comprehends the structure of formal languages, which not only enhances the interpretability of the Transformer-based ML systems but also marks a crucial step toward a deeper understanding of how ML systems process formal languages. Code and data are available at \url{https://github.com/Zhang-Yihao/Transfomer2DFA}.
\end{abstract}


\maketitle

\section{Introduction}
\label{chap:intro}
\begin{figure*}[]
    \centering
    \includegraphics[width=\linewidth]{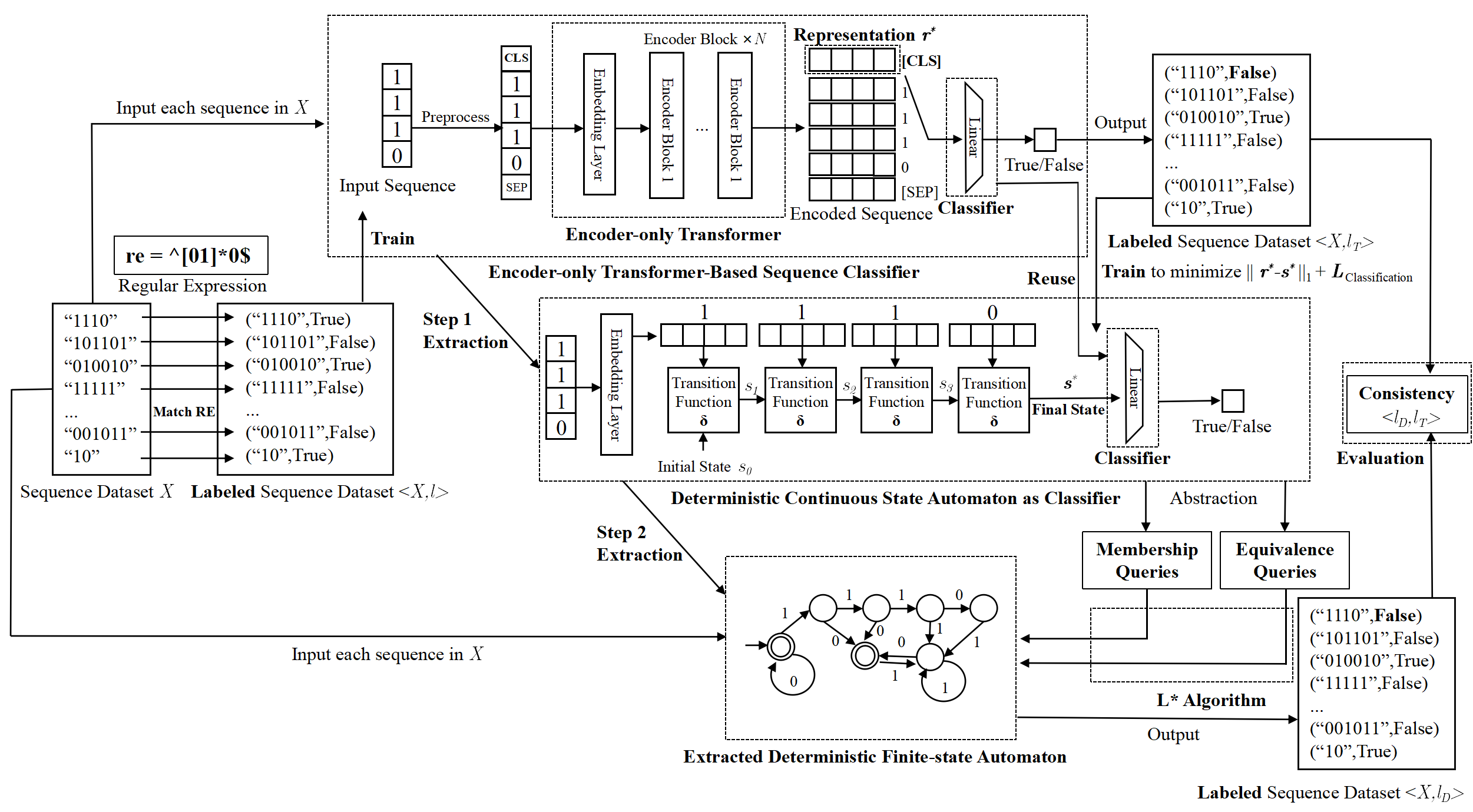}
    \caption{An illustration for overall extraction pipeline.}
    \label{fig:extract}
\end{figure*}
In the realm of machine learning (ML) systems, handling data structured in sequences was a fundamental challenge. Two main lines of model structures have been developed to tackle this problem. The first track of models, known as the Stateful Sequential Model (SSM)~\cite{gu2022efficiently}, represents an evolution of finite state machines adapted for ML systems. This group includes technologies such as Recurrent Neural Networks (RNNs)~\cite{medsker2001recurrent} and Long Short-Term Memory Networks (LSTMs)~\cite{sak2014long}. These models process data sequentially, mimicking the logical flow of traditional state machines, and are particularly effective in simulating decision-making processes similar to those found in finite-state automata, which are systems designed to manage a series of inputs to produce a specific output.

In contrast, the second line of models focuses on the broader contextual relationships within the data, rather than strictly adhering to the sequence of inputs. This approach is exemplified by the \textbf{Transformer} model~\cite{Attention}, which uses advanced mechanisms such as attention and positional encoding to interpret the significance of text segments within a broader context. Introduced in the seminal paper \textit{Attention is All You Need}~\cite{Attention}, Transformers have revolutionized natural language processing (NLP), demonstrating superior performance in tasks like machine translation, text summarization, and sentiment analysis~\cite{wang2019learning,floridi2020gpt}. Unlike SSMs, transformers do not rely on a traditional state-based architecture but instead highlight the interactions within the data sequences. Although initially considered devoid of state mechanisms, ongoing research suggests that transformers might implicitly involve state-like mechanisms~\cite {dao2024transformers}, yet this area remains under active investigation. In the transformer model family, encoder-only architectures like BERT~\cite{devlin2019bert} play a pivotal role in sequence classification tasks, akin to the function of state machine models in handling sequential data. 

Overall, both the two model types can effectively process language data and address various types of language tasks. Building upon these foundations, researchers are exploring whether the inherently stateless design of transformers can mimic traditional state machines to some extent~\cite{dao2024transformers}. Furthermore, the operational dynamics of transformers, especially in how they process and understand language, remain an active area of research~\cite{strobl2024formal}. These explorations can be further extended by using formal languages, particularly regular languages describable by automata, to analyze the operational states of these ML models. More specifically, recent studies~\cite{bhattamishra2020ability,Hahn_2020} indicate that transformers do not perform as well on tasks involving formal languages when compared to state-based models such as RNNs and LSTMs. This suggests a gap in transformers' ability to handle the rigid syntax required by formal languages, despite their proficiency with the fluid semantics of natural language. This discrepancy underscores the necessity for extensive research focused on interpretive studies and investigations into the internal mechanisms of transformers to bridge this gap. 

By extracting finite-state automata from these models and testing them with certain formal languages, researchers can assess if there is a form of equivalence between certain ML model architectures for sequential data processing and traditional state machines. This line of inquiry could potentially extend further into understanding how models manage state information and their internal operational dynamics, thus enhancing our knowledge to solve the issues we mentioned above. Efforts have been made to extract finite state automata from state space machine learning models and to test these extractions on certain formal languages, thereby assessing a form of equivalence with traditional state machine models~\cite{weiss2020extracting,xu2021extracting}. Similarly, conceptualizing transformers as deterministic finite automata might offer a higher level of interpretability and transparency, allowing researchers to better understand their operational dynamics in processing formal languages. However, despite this promising direction, the field lacks established methodologies for extracting automata from ML systems based on transformer architectures.

In this paper, we introduce a novel approach to enhancing the interpretability of encoder-only transformers in processing formal languages by extracting deterministic finite automata (DFAs) using representation-based abstraction combined with the \textit{L}$^*$ algorithm~\cite{weiss2020extracting} in a \textit{fully automated} way. Our methodology builds upon previous work in the area of automata extraction from RNNs and adapts it to the unique architecture of transformers, which traditionally do not exhibit clear state-based behavior typical of RNNs or finite state machines. Specifically, our approach begins by treating the transformer model as a black box, exploring its ability to simulate the behavior of a deterministic continuous state automaton (DCSA), a novel automaton architecture we proposed to support the extraction. By leveraging the internal representations produced by the transformer's encoder~\cite{zou2023representation}, we propose to simulate these representations as states of a hypothetical automaton. In this way, we can simulate the transformer as a stateful automaton thus leveraging existing methods to further extract finite-state automata from the simulated model. We evaluate the effectiveness of our extraction method through several metrics, primarily focusing on the accuracy of the extracted DFA in reproducing the behavior of the original transformer model across unseen test sequences. The overall illustration for our complete extraction pipeline is shown in Figure~\ref{fig:extract}.

In summary, our main contributions are as follows:
\begin{enumerate}
    \item We presented a novel, automated, and general framework for extracting deterministic finite-state automaton from en\-coder-only transformers. 
    \item We conduct abundant experiments on training BERT in formal languages, showcasing the effectiveness of our method while providing in-depth insights into the inner mechanisms of transformers.
    \item We analyze the results of extracted automatons to enhance the interpretability and transparency of the transformer models.
\end{enumerate}

This paper is organized as follows. In Section 2, we provide a comprehensive background on the relevant literature, including foundational concepts in transformer architectures and automata theory. Section 3 details our methodology for extracting deterministic finite automata from encoder-only transformers, outlining the technical approaches and the adaptation of the \textit{L}$^*$ algorithm. Section 4 describes the experimental setup, including the datasets used, the training of transformer models, and the criteria for evaluation. Section 5 presents the results of our experiments, providing quantitative and qualitative analysis to assess the effectiveness and accuracy of the extracted automata. Section 6 discusses related works. Finally, Section 7 concludes the paper with a summary of our contributions and a reflection on the impact of our research. 
\section{Preliminaries}
\label{chap:bg}

\textbf{Deterministic Finite Automaton.} A \textit{deterministic finite automaton} (DFA) is formally defined as a five-tuple $A = (\Sigma, S, s_0, F, \delta)$, where:
\begin{itemize}
    \item $\Sigma$ is a finite, nonempty set of input symbols, representing the alphabet,
    \item $S$ is a finite, nonempty set of states,
    \item $s_0 \in S$ is the initial state,
    \item $F \subseteq S$ is the set of accepting states,
    \item $\delta: S \times \Sigma \to S$ is the transition function.
\end{itemize}
A DFA serves as a binary acceptor for formal languages, particularly \textit{regular languages}, by iteratively applying the transition function $\delta$. The acceptance function $\Delta(\cdot): \Sigma^* \rightarrow \{0,1\}$ is derived from $A$ and can be recursively defined with a helper function $\Delta'$ such that $\Delta'(\varepsilon) = s_0$ and $\Delta'(w \cdot w_{-1}) = \delta(\Delta'(w), w_{-1})$ for $w \in \Sigma^*$ and $w_{-1} \in \Sigma$. Here, a string $x$ is accepted by $A$ (i.e., $\Delta(x) = 1$) if and only if $\Delta'(x) \in F$.

\textbf{Deterministic Continuous State Automaton.} To support our extraction pipeline, we propose the \textit{deterministic continuous state automaton} (DCSA) architecture which is defined similarly to the DFA but operates on a continuous state space. An $n$-dimen\-sional DCSA is defined as a five-tuple $S = (\Sigma, S, s_0, F, \delta)$, where:
\begin{itemize}
    \item $\Sigma$ is a finite, non-empty set of input symbols (alphabet),
    \item $S \subseteq \mathbb{R}^n$ represents the continuous states,
    \item $s_0 \in \mathbb{R}^n$ is the initial state,
    \item $F: S \rightarrow \{0,1\}$ is a function that accepts or rejects a state based on certain criteria,
    \item $\delta: S \times \Sigma \rightarrow S$ is the transition function.
\end{itemize}
The acceptance function $\Delta(\cdot): \Sigma^* \rightarrow \{0,1\}$ for a DCSA can also be recursively defined with a helper function $\Delta'$ such that $\Delta'(\varepsilon) = s_0$ and $\Delta'(w \cdot w_{-1}) = \delta(\Delta'(w), w_{-1})$ for $w \in \Sigma^*$ and $w_{-1} \in \Sigma$. The acceptance of a string $x$ in this context is determined by $\Delta(x) = F(\Delta'(x))$. The DCSA framework can represent various existing continuous-state recurrent models, including Weighted Finite Automata (WFA)~\cite{droste2009handbook}, Recurrent Neural Networks (RNN)~\cite{medsker2001recurrent}, Long-Short Term Memory (LSTM) based RNNs~\cite{sak2014long}, and Gated Recurrent Units (GRU)~\cite{chung2014empirical}. Utilizing this architecture, we can analyze the similarities between ML-based systems and traditional DFAs in processing formal languages.

\textbf{Encoder-Only Transformers.} Encoder-only trans\-formers, c\-ommonly referred to as BERT-like models~\cite{devlin2019bert}, are transformer models that exclusively employ encoder blocks in their architecture. These models are primarily designed for sentence comprehension, particularly for string classification in this context, while decoder-only transformers focus on predicting the next token. Consequently, we utilize encoder-only transformers as binary acceptors by training them on formal languages as sentence classifiers. Each input sentence is tokenized and appended with special tokens \texttt{[CLS]} and \texttt{[SEP]} at the beginning and end, respectively, to denote the absence of inherent semantic meaning. After undergoing $n$ layers of cross self-attention calculations, the encoded embedding for the \texttt{[CLS]} token serves as the representative encoded information for the entire sentence at the initial stage. Formally, for an input sequence $X = (x_1,\cdots,x_m)$ and an $n$-layer encoder-only transformer $M$, the output function of the transformer can be expressed as $M(X) = F(f_n(\cdots f_1(E(X))) \cdots)$, where $|X| = m$, $E: \Sigma^m \rightarrow \mathbb{R}^{m \times n}$ is the embedding function mapping a sequence to a series of real vectors, $f_i : \mathbb{R}^{m \times n} \rightarrow \mathbb{R}^{m\times n}$ represents the $i$-th encoder block performing self-attention and layer normalization, and the output of $f_i$ is also referred to as the \textbf{hidden state}. Finally, $F:\mathbb{R}^{m \times n} \rightarrow \{0,1\}$ denotes the classification function mapping the final representation to acceptance or rejection, typically implemented through machine learning classification algorithms on the representation of the last layer. Hereby, we formally define the \textbf{representation} corresponding to a certain input token sequence $X$ as
\begin{equation}
    \text{Rep}(X) := f_n(\cdots f_1(E(X)) \cdots )\texttt{[CLS]},
\end{equation} indicating the output value of the last encoder block at the \texttt{[CLS]} token.

\textbf{Regular Language and Regular Expressions.} 
Regular expressions are a powerful and concise way to describe patterns within strings, using a variety of symbols to specify complex string structures. Basic symbols in regular expressions include literals (specific characters), the asterisk (*) for zero or more repetitions, the plus sign (+) for one or more repetitions, the question mark (?) for zero or one occurrence, and the pipe (|) for alternatives. Parentheses can group parts of the expression to enforce precedence. These symbols enable the construction of intricate patterns that can represent sequences of characters, choices between alternatives, and repetitions. Specifically, every regular expression can be translated into an equivalent DFA~\cite{yu1997regular}, which is a state machine designed to recognize the same set of strings described by the regular expression. This equivalence ensures that any language definable by a regular expression is also recognizable by a DFA, and vice versa.

\textbf{The \textit{L}$^*$ Algorithm.}
The \textit{L}$^*$ Algorithm~\cite{angluin1987learning} is a well-established method for the automatic learning of automata. It operates through interactive queries to a teacher regarding the behavior of the target automaton, involving both membership and equivalence queries. Membership queries determine if a specific string is accepted by the automaton, while equivalence queries assess whether the current hypothesis automaton behaves identically to the target. When discrepancies arise, the teacher provides a counterexample that is used to refine the hypothesis. This process of iterative refinement continues until the algorithm successfully identifies the correct automaton configuration, showcasing the efficiency of \textit{L}$^*$ in contexts requiring precise modeling.

Building on this foundational approach, Weiss et al.~\cite{weiss2020extracting} introduced a novel methodology for learning automata from Recurrent Neural Networks (RNNs) inspired by the \textit{L}$^*$ Algorithm. We denote the extracted automaton as $A$. Their technique directly employs the RNN classifier to conduct membership queries. For equivalence queries, it leverages the abstraction method proposed by Omlin and Giles~\cite{Omlin1994ConstructingDF} to abstract $A^{R,p}$ from the original RNN $R$, based on a partition $p$ of the state space $S$. The subsequent step to refine the extracted automata $A$ is performed by comparing the behavior of the given DFA $A$ with the abstraction $A^{R,p}$. Any disagreement is either used as a counterexample or to further refine the partition $p$, facilitating iterative adjustments until $A$ aligns with $A^{R,p}$, thus validating the extracted DFA.

It is important to note that this approach is not strictly limited to RNNs. In fact, it can be extended to all Deterministic Continuous State Automata (DCSAs). By approximating the transition function $\delta$ in a DCSA using an RNN layer, any DCSA can effectively be treated as equivalent to an RNN. This broader applicability defines our proposed paradigm for extracting automata from DCSAs, a methodology whose effectiveness is corroborated by the findings in~\cite{weiss2020extracting}.

\textbf{Representation Engineering.}
Representation Engineering, introduced by Zou et al.~\cite{zou2023representation}, offers a novel top-down perspective to understand the behavior of transformers. This approach focuses on the representations derived from the embeddings of middle layers within the transformer. These representations are shown to be highly indicative of the transformer's overall behavior. For enhanced explainability of our extraction method, we utilize the representation space as the first-class object for our extracting approach, akin to the state space leveraged in the RNN extraction process. Further details on how this representation space is employed in our extraction methodology will be discussed subsequently.
\section{Methods}
\label{chap:mtd}

Our approach employs a two-step DFA extraction process designed to elucidate the inner workings of encoder-only transformers. This can be briefly organized as below:

\textbf{Step 1 (Transformer $\rightarrow$ DCSA).}
We simulate the internal dynamics of encoder-only transformers as deterministic continuous-state automata to approximate their state transitions during sequence processing.

\textbf{Step 2 (DCSA $\rightarrow$ DFA).}
Utilizing the \textit{L}$^*$ algorithm, we iteratively refine and extract a deterministic finite automaton that encapsulates the precise patterns and decision-making processes learned by the transformer.

In the rest of this section, we provide details for this two-step extraction pipeline. 
\subsection{Motivations}
Having established a paradigm for extracting DFAs from DCSAs using the method described in~\cite{weiss2020extracting} as referred to in Section~\ref{chap:bg}, we consider transforming an encoder transformer into a DCSA and applying the extended version of the algorithm initially designed for RNNs. However, the unique characteristics of transformers, which process sequential data concurrently without explicit hidden states or transformation functions for each token, pose significant challenges. Moreover, a fundamental barrier to directly applying existing extraction algorithms is the inherent non-determinism of transformer transformations. Transformers utilize positional embedding to incorporate token positional information during the embedding process. This mechanism inherently introduces non-determinism, as the initial embedding of a token varies depending on its positional embedding. Without these embeddings, transformers would struggle to recognize formal language classes because the semantics of a token in formal languages are critically dependent on its position in the sequence.

These observations highlight two primary challenges for our extraction algorithm: \textit{firstly}, identifying an appropriate analog to the state space used in existing RNN extraction approaches; and \textit{secondly}, addressing the non-deterministic nature of transformers, since the ultimate output of our algorithm, the DFA, must be deterministic.

To address these challenges, we propose a two-step extraction algorithm tailored for encoder-only transformers, aimed at precisely capturing what the transformer has learned embodied in the form of DFA. Our presentation of this algorithm begins by detailing solutions to the identified issues, followed by an illustration of how these solutions are integrated into our methodology.

To address the first challenge of identifying an object akin to the state space utilized in RNNs, we opt for the \textbf{representation space} to fulfill the role traditionally played by state space. Formally, representation space $S\subset \mathbb{R}^n$ is defined as $\{\text{Rep}(X)|X\in \Sigma^*\}$. The representation vector, generated during sentence processing and specifically defined in Section~\ref{chap:bg}, is the embedding vector produced in the last layer of the special token \texttt{[CLS]}. This vector is selected as the analog to the state in RNNs for several reasons:

Firstly, the representation vector mirrors the role in the DCSA we previously defined; with each new token appended to the sequence, this vector undergoes modifications that reflect the nature of the token, albeit influenced by the context and its position. Treating the classifier function $F$ in the transformer as the acceptor function in a DCSA, the final representation vector, once all tokens have been processed, determines the classification outcome. This is akin to how the state vector functions in an RNN.

Secondly, prior studies such as Representation Engineering~\cite{zou2023representation} and foundational work on BERT~\cite{devlin2019bert} demonstrate that the defined representation vector effectively captures the transformer's operational dynamics and offers a comprehensive reflection of the internal state transformations. This underlines its potential to represent the overall network state.

It is important to note, however, that there is no real ``representation transition'' with each new token since the sequence is processed concurrently in transformers. Thus, the concept of state transitions here is more of a ``simulation'' rather than a true depiction of how transformers understand formal languages. This adaptation is necessary due to the lack of discrete states in transformers, yet we must designate an equivalent to facilitate the deterministic automaton abstraction.

To address the second challenge, namely the inherent nondeterministic nature of transformers which complicates the direct extraction of behavior into a DFA, we propose an abstraction using a DCSA based on the representation space. This DCSA approximates the transformer's computations and serves as the basis for applying the subsequential DFA extraction, utilizing the \textit{L}$^*$ algorithm. If the transformer, the abstracted DCSA, and the subsequently extracted DFA demonstrate high consistency in classification results across a formal language dataset, we can confidently assert that the extracted DFA accurately represents the patterns learned by the original transformer.

\subsection{Algorithm Design}
To delve deeper into the details, we formally outline our approach as follows: Consider a formal language $L$ and another sequence dataset $X \subset \Sigma^*$, which serves as the language dataset. For each sequence $x \in X$, we define the corresponding label function $l$ as the given classification function $F: X \rightarrow \{0,1\}$.

In the original training scenario, we use the characteristic function of the formal language $L$, denoted as $\mathbf{1}_{x \in L}$, as the label function for the ground truth. We then consider a transformer model, $T: X \rightarrow \{0,1\}$, as another classification function over the dataset $X$. Consequently, we generate a labeled dataset $(X,T)$, which contains the same instances as $X$ but uses the label function $T$.

The consistency rate between two label functions $l$ and $l'$ on the same dataset $X$ is defined as 
\begin{equation}\begin{aligned}
C(l,l') = \frac{|\{x \in X \mid l(x) = l'(x)\}|}{|X|},
\end{aligned}\end{equation}
which represents the proportion of samples where both label functions agree. The transformer model, parameterized by $T_{\theta}$, is trained to maximize this consistency rate $C(\mathbf{1}_{x \in L}, T_{\theta})$ with respect to the ground truth. We define $\mathcal{L}_C$ as the loss function for the training process of the transformer model, which is ordinarily the sum of classification cross-entropy loss on sequences in $X$, hereby the optimization goal can be formally described as 
\begin{equation}\begin{aligned}
\min_\theta \mathcal{L}_C = &\sum_{x \in X} -(\mathbf{1}_{x \in L}(x)\log \mathbb{P}[T_\theta(x) = 1] \\ &+ (1-\mathbf{1}_{x \in L}(x))\log \mathbb{P}[T_\theta(x) = 0]).
\end{aligned}\end{equation}

We say a transformer model learns some pattern from a labeled dataset $(X, l)$ if the trained model $T$ achieves a consistency rate $C(\mathbf{1}_{x \in L}, T) > 1/2$ stably. Nevertheless, situations may arise where the transformer cannot adequately approximate the original data distribution, leading to a low consistency rate, potentially around $1/2$. However, this does not impact the ultimate goal of understanding the model's learned behavior, provided the consistency rate between the final extracted DFA and the original model is significantly higher than random guessing.

To effectively abstract a transformer model into a deterministic continuous system, we designate the DCSA with the symbol $D$. This system functions as a classification mechanism over the dataset $X$. To enhance its efficacy and align it with the deep, internal features learned by the transformer—especially in terms of state representation—we incorporate a representation generation function, $\text{Rep}_\theta(\cdot)$, derived from the transformer $T_{\theta}$ as demonstrated in Section~\ref{chap:bg}. This function is crucial for guiding the optimization of the DCSA. Practically, we train the DCSA to mimic the distribution of the representation at its final state. This approach aligns the internal features between the DCSA and the transformer model, ensuring consistency in their feature representations. Additionally, for the final state acceptor function $F$, previously defined for the DCSA, we opt to \textbf{reuse} the classification mechanism employed in the transformer model $T_{\theta}$ with the same layer architecture and weights. This approach ensures a high level of consistency between the two models.

Parameters of DCSA are optimized targeting the two primary objectives below:

    \textbf{(1) Classification Consistency Rate,} denoted as consistency rate between label functions derived from the two models $C(T_{\theta}, D_{\gamma})$. This metric should be maximized to ensure that the classification outcomes from the DCSA align closely with those from the transformer. This can be achieved by minimizing the classification cross-entropy loss, expressed as:
    \begin{equation}\begin{aligned}\mathcal{L}_D &= \sum_{x \in X} -(T_{\theta}(x)\log \mathbb{P}[D_{\gamma}(x) = 1] \\ &+ (1-T_{\theta}(x))\log \mathbb{P}[D_{\gamma}(x) = 0]);\end{aligned}\end{equation}
    
    \textbf{(2) Representation Distance,} which measures the disparity between the internal state of the DCSA and the final representation of the transformer when processing the same token sequence. Formally, it is defined as
    \begin{equation}\begin{aligned}\mathcal{L}_{\text{Rep}} = \sum_{x\in X} \alpha||\text{Rep}(x) - \text{State}(x)||_{L^1},\end{aligned}\end{equation}
    where $\text{State}(x)$ represents the final state of the DCSA when processing $x$, as $\Delta'(x)$ defined in Section~\ref{chap:bg}, and $\alpha$ is a coefficient to balance the two losses. This metric is aimed to be minimized to align the internal consistency level with the transformer.
    
Assuming that the DCSA is parameterized by $\gamma$, the optimization objective for $D_{\gamma}$, with the given representation function $\text{Rep}$, is framed as achieving the best balance between these two goals, formally
\begin{equation}\begin{aligned}
\min_{\gamma} \mathcal{L}_D + \mathcal{L}_{\text{Rep}}.
\end{aligned}\end{equation}

For implementation issues, this minimization is in fact implemented by alternatively optimizing both losses. This approach aims to closely align the DCSA's classification outputs with those of the original label function, leveraging both the transformer's learned representations and the inherent structure of the data. Details of the optimization will be provided in Section~\ref{chap:set}.

\begin{algorithm}[t]
\caption{Extracting DFA from Encoding-Only Transformers}\label{alg}
\SetKwFunction{ext}{ExtractDfaFromDcsa}
\KwData{Regular Language $L$, Sequence Dataset $X$, Encoder-Only Transformer $T_{\theta}$, DCSA $D_{\gamma}$, Epoch number for training transformer $N_T$, Epoch number for training DCSA $N_D$}
\KwResult{DFA $A$}
\textbf{def $\mathcal{L}_C(\theta) = $} \Comment{Loss for transformer classification task} \\ \quad \quad $\sum_{x \in X} -(\mathbf{1}_{x \in L}(x)\log \mathbb{P}[T_{\theta}(x) = 1] + (1-\mathbf{1}_{x \in L}(x))\log \mathbb{P}[T_{\theta}(x) = 0])$\\
\textbf{def $\mathcal{L}_D(\theta,\gamma) = $} \Comment{Loss for disagreement between transformer and DCSA} \\ \quad \quad $\sum_{x \in X} -(T_{\theta}(x)\log \mathbb{P}[D_{\gamma}(x) = 1] + (1-T_{\theta}(x))\log \mathbb{P}[D_{\gamma}(x) = 0])$\\
\textbf{def }$\mathcal{L}_{\text{Rep}} = \sum_{x\in X} ||\text{Rep}(x) - \Delta'(x)||_{L^1}$ \Comment{Penalize inconsistent between DCSA state and representation}\\
\For{$N_T$ epochs}{
Update $\theta$ by descending $\nabla_\theta \mathcal{L}_C$ \Comment{Optimize $\theta = \arg \min_\theta \mathcal{L}_C$}\\
}
\For{$N_D$ epochs}{
Update $\gamma$ by descending $\nabla_\gamma \mathcal{L}_D$\\
Update $\gamma$ by descending $\nabla_\gamma \mathcal{L}_\text{Rep}$ \Comment{Optimize $\gamma = \arg \min_\gamma \mathcal{L}_D+\mathcal{L}_{\text{Rep}}$}\\
}
$\Return \ \ext(D_{\gamma})$\Comment{Using \textit{L}$^*$ Algorithm}\\
\end{algorithm}

Hereby the overall pipeline can be concluded by finally applying \textit{L}$^*$ algorithm to extract DFA $A$ from DCSA, leveraging the extended version of an algorithm proposed in~\cite{weiss2020extracting}. We finally summarize our two-step extracting algorithm in Algorithm~\ref{alg}. The overall performance of extraction is measured by the consistent rate between the extracted DFA and original model $T$, formally $C(l_A, l_{T_{\theta}})$. This target is expected to be close to 1, which could provide certification for successfully extracting the \textit{real} pattern transformer learned initially.

\section{Experimental Settings}
\label{chap:set}
\label{expset}
\textbf{Dataset.} It's noteworthy that the formal language class that DFA could recognize is exactly regular language. Considering that our research is limited to the field of extracting DFA, so all formal language is chosen as regular languages. We mainly evaluate our model on three sets of regular languages:
\begin{enumerate}
    \item \textit{Tomita Languages}~\cite{tomita1982dynamic}. They serve as a universal benchmark for tasks involving formal languages, particularly tho\-se related to DFAs. They are well-known and widely used in the field for evaluating automata extraction frameworks.
    \item  \textit{Mod $n$ Langugages}. This category comprises regular languages that represent binary numbers. In these languages, the strings that are divisible by $n$ are accepted. This set is particularly useful for evaluating the capability of DFAs to recognize patterns related to modular arithmetic in binary representations.
    \item \textit{Transformer Formal Languages}~\cite{bhattamishra2020ability}. This dataset includes a diverse set of regular languages, allowing for the extension of our evaluation to accommodate larger alphabet sizes and more complex patterns. This resource provides a robust framework for testing and analyzing the capabilities of our model in handling a wider variety of regular languages, ensuring a comprehensive assessment of its performance.
\end{enumerate} 
We create a sequence dataset for each regular expression. For regular expression $r$, we denote its corresponding regular language as $L = \{x|x\text{ matches }r \land x \in \Sigma^*\}$. Basically the scale of dataset $X$ is in the order of $10^3$. Sequence dataset $X$ consists of randomly created sequences with various lengths. To ensure the effectiveness of our training and testing process, the sequence dataset is constructed in a balanced way, formally $|X \cap L|/|X| \approx 1/2.$ 

\textbf{Structure of Models.}
For encoder-only transformers to be train\-ed on formal language datasets, we refer to several representative model structures for encoder-only transformer architectures as po\-inted out by Huggingface Hub~\cite{huggingfacecourse}. We choose three representative transformers: ALBERT~\cite{lan2020albert}, BERT~\cite{devlin2019bert}, and DistilBERT~\cite{sanh2020distilbert}. For DCSA, as the middle product of extraction, we select several well-known implementations in the machine learning field, including RNN~\cite{medsker2001recurrent}, LSTM~\cite{sak2014long}, and GRU~\cite{chung2014empirical}. All mentioned models are implemented using the official implementations provided by \texttt{PyTorch}~\cite{NEURIPS2019_9015} and \texttt{Transformers}~\cite{wolf-etal-2020-transformers}. For the classifier for DCSA and transformers, they use the same single linear layer with the same weights, as mentioned in Section~\ref{chap:mtd}. This layer maps the input representation/state to $\mathbb{R}^2$, representing the confidence levels for whether the original sequence belongs to $L$ or does not belong to $L$. To unify the internal features and representations in the transformer and DCSA, we use the same 32-dimensional real vector to encode tokens in sequences. This approach allows us to demonstrate the intermediate states and the representations in a unified way, making the encoded information transferable between models and layers. Other parameters are chosen by default and provided in the library of \texttt{Transformers}.

\textbf{Training Arguments and Environments.}
We train the transformers using the trainer class provided in the \texttt{Transformers} library, utilizing the AdamW optimizer~\cite{loshchilov2019decoupled} to optimize the parameters. The training consists of 200 epochs of gradient descent with a learning rate of $5 \times 10^{-4}$. For training DCSA, we use the same optimizer and optimization steps but with a learning rate of $1 \times 10^{-3}$. For this part, the parameters of classifier is frozen after copying weights from the transformer. The batch size is set to 1 for both DCSA and transformer training. All experiments are conducted on a server with an Intel(R) Xeon(R) Gold 6430 CPU running at 2.60 GHz and an NVIDIA GeForce RTX 4090 GPU with CUDA version 12.4.

\section{Exprerimental Results and Analysis}
We evaluate our extraction method and its possible downstream application by answering the following research questions (RQs):
\begin{itemize}
    \item \textbf{RQ1 (Effectiveness).} How effective is our extraction method?
    \item \textbf{RQ2 (Generalization).} Is our method suitable for different encoder-only transformer architectures and DCSAs?
    \item \textbf{RQ3 (Explainability).} How can our method explain the behaviors of encoder-only transformers in formal languages?
    \item \textbf{RQ4 (Ablation).} Is it necessary to introduce the representation-based technique into this extraction process?
\end{itemize}
\subsection{Effectiveness (RQ1)}
\label{effectiveness}
\begin{center}
\begin{tcolorbox}[colback = black!10,width=\linewidth]
    {How \textit{effective} is our extraction method?}
\end{tcolorbox}
\end{center}

\begin{table}[]
    \centering
\begin{tabular}{c|c|c|c|c}
    \toprule
    Grammar & Learnability & $C(L,T)$ & $C(T,D)$ & $C(T,A) (\uparrow)$ \\
    \midrule
    (aa)$^*$ &\checkmark & 1.0000& 1.0000& \textbf{1.0000}\\
    (abab)$^*$ & \checkmark& 1.0000& 1.0000& \textbf{1.0000}\\
    Parity &\usym{2613} & 0.5013& 0.4994& \textbf{0.9963}\\
    $\mathcal{D}_2$ &\checkmark & 0.9824 & 0.9899& \textbf{0.9899}\\
    $\mathcal{D}_4$ &\checkmark & 0.9557 & 0.9557 & \textbf{0.9536}\\
    \midrule
    Tomita 1 & \checkmark & 1.0000 & 1.0000& \textbf{1.0000}\\
    Tomita 2 & \checkmark & 1.0000 & 1.0000& \textbf{1.0000}\\
    Tomita 3 & \checkmark & 0.9217 & 0.8995 & \textbf{0.9731}  \\
    Tomita 4 & \checkmark & 1.0000 & 1.0000& \textbf{1.0000}\\
    Tomita 5 & \usym{2613} & 0.5313 & 0.7791 & \textit{0.7569} \\
    Tomita 6 & \usym{2613} & 0.5017 & 0.5291 & \textit{0.7603}\\
    Tomita 7 & \checkmark & 1.0000& 1.0000& \textbf{1.0000}\\
    \midrule
    Mod 2& \checkmark& 1.0000& 1.0000& \textbf{1.0000}\\
    Mod 3& \usym{2613} & 0.4783 & 0.6534 & \textit{0.6911}\\
    Mod 4 & \checkmark& 1.0000& 1.0000& \textbf{1.0000}\\
    Mod 5&\usym{2613} & 0.5144 & 0.7033 & \textit{0.7217}\\
    \bottomrule
    
\end{tabular}
    \caption{Evaluation results on extracting DFA from BERT trained on various regular languages, utilizing an RNN for the DCSA throughout the extraction process. Values in \textbf{bold} indicate outcomes that closely match the original model, demonstrating accurate extractions. Values in \textit{italic} denote results that, while not perfectly fitting the original model, are still considerably better than random selection, thereby offering valuable insights into the operational dynamics of transformers.
}
    \label{tab:tomita}
    \vspace{-1cm}
\end{table}

\begin{figure}[]
    \centering
    \includegraphics[width = 0.4\linewidth]{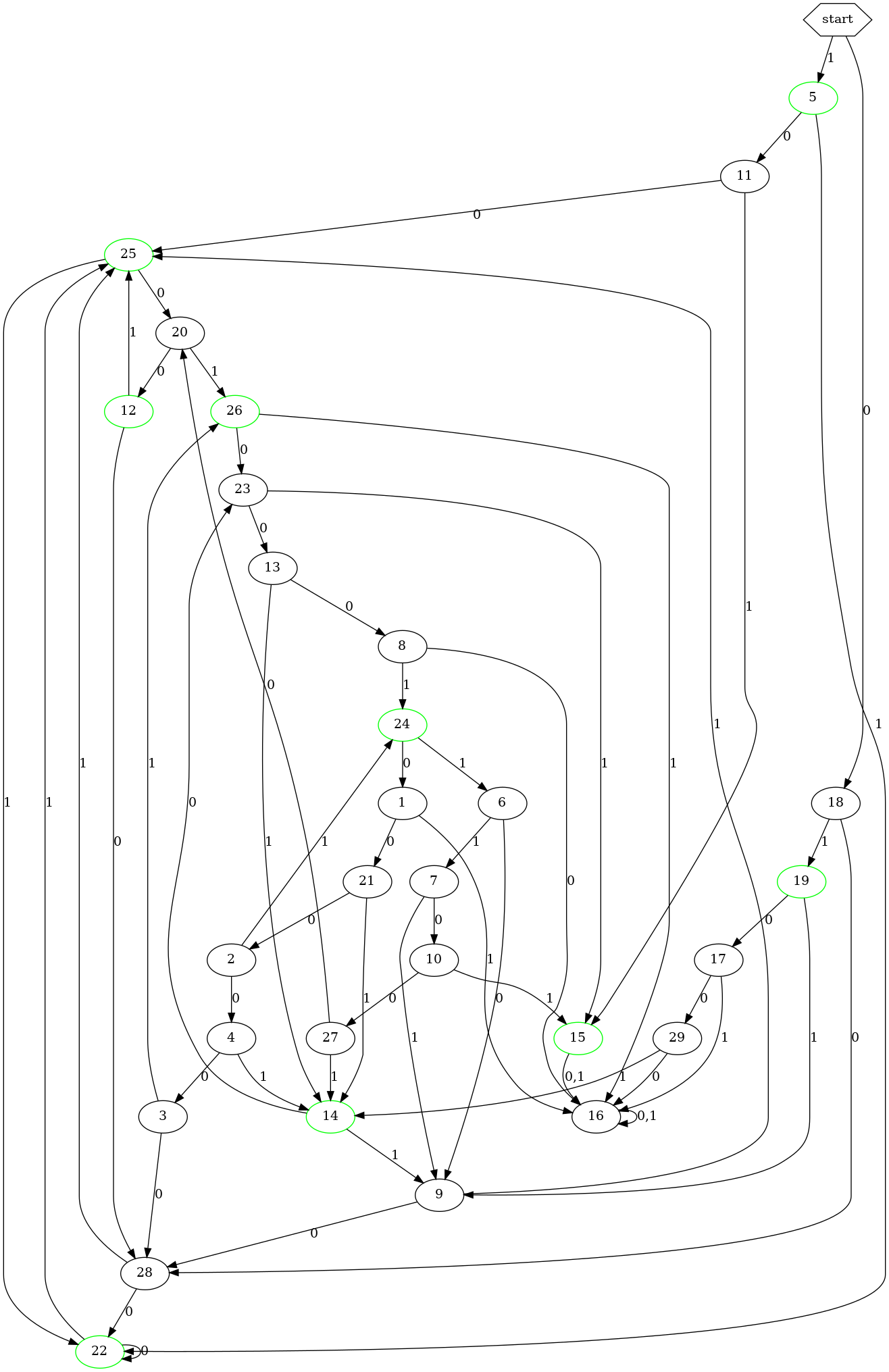}
    \caption{Extracted automaton from transformer trained on Tomita 3 language, which is exactly the complement of $((0|1)^{*}0)^{*}1(11)^{*}(0(0|1)^{*}1)^{*}0(00)^{*}(1(0|1)^{*})^{*}$ over the alphabet $\{0,1\}$. The extraction result is basically correct, with a consistent rate over 97\%.}
    \label{fig:automaton-tomita-3}
\end{figure}

\begin{figure}[]
    \centering
    \includegraphics[width=0.2\linewidth]{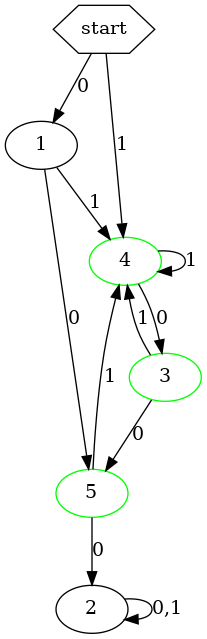}
    \caption{Extracted automaton from transformer trained on Tomita 4 language, which contains all words in $\{0,1\}^*$ containing no "000". The extraction result is completely correct.}
    \label{fig:automaton-tomita-4}
\end{figure}

\begin{figure}[t]
    \centering
    \includegraphics[width=0.3\linewidth]{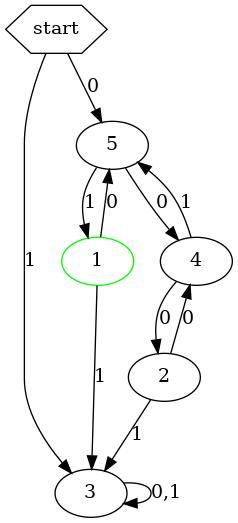}
    \caption{Extracted automaton from transformer trained on $\mathcal{D}_4$ language. $\mathcal{D}_n$ is defined recursively as a regular language, where $\mathcal{D}_0 = \varepsilon$ and $\mathcal{D}_n = (0 \mathcal{D}_{n-1} 1)^*$. The extraction result is completely correct.}
    \label{fig:automaton-d4}
\end{figure}

The experimental results assessing our method across the designated datasets are presented in Table~\ref{tab:tomita}. All values are attained under the basic model structure, which is BERT for the transformer and RNN for DCSA. To gauge the efficacy of our approach, particularly since the task centers around extraction, we employ the \textit{consistent rate} as a metric for evaluation. Regardless of the transformer's generalization capabilities on the formal language dataset, we can assess the congruence between the final DFA and the original model to determine the precision of our extraction method. We denote the labeling functions corresponding to the regular expression, transformer, DCSA, and DFA as $L, T, D,$ and $A$, respectively. These enable us to compute various consistent rates between models to evaluate the effectiveness of the extraction process across different stages.

In the table, three primary consistent rates serve as indicators: the consistent rate between the original regular language and the transformer denoted as $C(L, T)$, which serves as a measure of the transformer's learnability for the corresponding language. It is no\-ted that certain formal languages, as identified in previous studies \cite{bhattamishra2020ability,Hahn_2020}, like Tomita 5 and Tomita 6, may not be learnable by the transformer. In these cases, a lower consistent rate is expected; however, our method can still discern some patterns through the extracted DFA. Most of the experimental results are consistent with the prior study on transformers for formal languages. Additionally, we report on the consistent rate between the provided DCSA and the trained transformer as $C(T, D)$, which indicates the accuracy of the initial extraction step. Although the final DFA is derived not directly from the transformer but rather from the DCSA, this metric can reflect how well the final DFA has captured the learned patterns.

Lastly, the most crucial evaluation metric for the overall process is presented: the consistent rate between the source transformer $T$ and the target DFA $A$, denoted as $C(T, A)$. Despite potential unlearnability issues with the original model, the DFA provides insights into what was actually learned during training, serving as a critical tool for model explanation. This metric is vital as it underscores how accurately the final DFA approximates the behavior of the original model.

To the best of our knowledge, our proposed method is the first to extract automata from transformer architectures, thereby lacking direct baselines for comparison. To discuss the effectiveness of our method, we categorize the regular languages into two types: \textit{learnable} and \textit{unlearnable} languages for transformers.

For learnable languages, where $C(L,T) \approx 1$, we observe that most extracted DFAs exhibit near-perfect consistency with the original transformers, achieving 100\% accuracy, especially when the transformers have flawlessly learned the patterns encoded in the given regular expressions. We illustrate two successful examples of such extractions in Figure~\ref{fig:automaton-tomita-4} and Figure~\ref{fig:automaton-d4}, where the automata accurately reflect the precise learning of the transformers for the specified regular languages.

For languages that are learnable but not perfectly captured by the trained transformer, the extraction accuracy remains high, arou\-nd 90\%, though not perfect. This slight discrepancy arises because the complexities learned by the transformer may be too intricate for a simple DFA to fully encapsulate. An example is provided in Figure~\ref{fig:automaton-tomita-3}, showing the extraction results for a transformer not fully trained on the Tomita 3 language. Despite not achieving perfect consistency, the high consistent rate still attests to the method's effectiveness.

For the \textit{unlearnable} languages, the extraction results hinge on the transformer's learning of the given pattern. Due to the limited expressiveness of DFAs, they may not capture all unexpected behaviors of transformers comprehensively. However, in most instances, the consistent rate between the DFA and the transformer significantly exceeds random choice, indicating $C(T,A) >> 1/2$, even when the transformer poorly learns the language. These results demonstrate that our method effectively identifies specific patterns learned by the transformers and can elucidate certain behaviors of trained transformers, ensuring the method's efficacy under various conditions.

\begin{center}
\begin{tcolorbox}[colback = white!100,width=\linewidth]
    {\textbf{Answer to RQ1:} Our extraction method has demonstrated robust effectiveness, as evidenced by the experimental results detailed in Table~\ref{tab:tomita}. Employing the \textit{consistent rate} as a metric, the method achieves impressive congruence between the extracted DFA and the original transformer models, with specific highlights including: (1) \textit{Overall High Consistency}-It achieves near-perfect consistent rates ($C(T,A) \approx 1.0$) for learnable languages, confirming the method's precision in capturing the patterns learned by the transformers; (2) \textit{Effective Across Various Scenarios}-For learnable languages not perfectly captured by the transformer, consistency still remains high (approximately 90\% on average), showcasing the method's capability to approximate complex patterns. (3) \textit{Resilient in Unlearnable Contexts}-Even in cases where languages are deemed unlearnable for transformers, the extracted DFA still significantly outperforms random choice, demonstrating the method's ability to discern underlying patterns despite the limitations of transformer and DFA.}
\end{tcolorbox}
\end{center}

\subsection{Generalization (RQ2)}
\begin{center}
\begin{tcolorbox}[colback = black!10,width=\linewidth]
    {Is our method suitable for \textit{different} encoder-only transformer architectures and DCSAs?}
\end{tcolorbox}
\end{center}

We conducted a series of experiments utilizing encoder-only transformers and various implementations of DCSAs across four representative regular languages: Tomita 1, 3, 5, and 7. These languages included two that are fully learnable by transformers (Tomi\-ta 1 and 7), one that is learnable but challenging to approximate accurately (Tomi\-ta 3), and one that is fundamentally unlearnable (Tomita 5). For our DCSA configurations, we selected LSTM and GRU models to explore the impact of different DCSA choices when extracting from a trained BERT model. Additionally, we employed ALBERT and DistilBERT as variations of BERT to assess the generalizability of our method. The experimental outcomes are detailed in Table~\ref{tab:generalize}.

\begin{table}[]
    \centering
\resizebox{\linewidth}{!}{\begin{tabular}{c|c|c|c|c|c}
    \toprule
    Grammar & BERT+LSTM & BERT+GRU & ALBERT+RNN & DistilBERT+RNN \\
    \midrule
    Tomita 1 & 1.0000 & 1.0000 & 1.0000& 1.0000\\
    Tomita 3 & 0.9305 & 0.9517 & 0.8887 & 0.8731  \\
    Tomita 5 & 0.7539 & 0.6813 & 0.7291 & 0.6569 \\
    Tomita 7 & 1.0000 & 1.0000& 1.0000& 1.0000\\
    \bottomrule
    
\end{tabular}}
    \caption{Evaluation results of $C(T,A)$ on extracting DFA on various encoder-only transformer architectures trained on representative Tomita regular languages. We use RNN as the DCSA leveraged in the overall extraction process. The values in bold represent the results of fitting the original model basically correctly.}
    \vspace{-1cm}
    \label{tab:generalize}
\end{table}

The outcomes of our experiments demonstrate the robust stability of our method across various structural configurations. As indicated in the table, the performance of our method across different combinations of transformers and DCSAs shows minimal variation. For the fully learnable languages, the results consistently reached 1.0, indicating perfect extraction. For Tomita 3 and Tomita 5, the results exhibited variations within a narrow range of 0.0786 and 0.0970, respectively, both below 0.1, underscoring the stability of our method. Additionally, the standard deviations for these variations are 0.0315 and 0.0382, respectively, which are relatively minor.

\begin{center}
\begin{tcolorbox}[colback = white!100,width=\linewidth]
    {\textbf{Answer to RQ2:} Our extraction method is suitable for different encoder-only transformer architectures and DCSAs. The experimental results consistently demonstrate high stability and minimal variation in performance across various configurations (with narrow ranges 0, 0.0786, 0.0970, and 0 for four given languages), affirming the method's adaptability and effectiveness for different systems.}
\end{tcolorbox}
\end{center}

\subsection{Explainability (RQ3)} 
\begin{center}
\begin{tcolorbox}[colback = black!10,width=\linewidth]
    {How can our method \textit{explain} the behaviors of encoder-only transformers on formal languages?}
\end{tcolorbox}
\end{center}

\begin{figure}[]
    \centering
    \includegraphics[width = 0.4\linewidth]{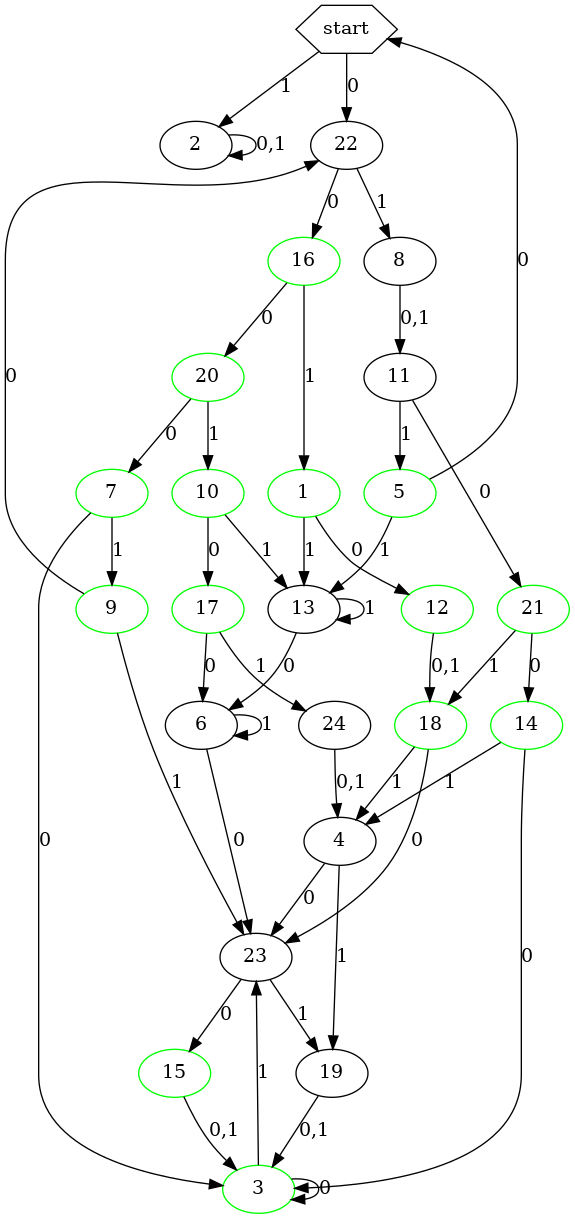}
    \caption{Extracted automaton from transformer trained on Mod 3 language, which contains all binary numbers that can be divided by 3.}
    \label{fig:automaton-mod-3}
\end{figure}

According to previous studies \cite{wei2024weighted,hong2022adaax}, extracting automata is considered a reliable method for elucidating the behaviors of neural networks, which are often characterized by their opaque, "black-box" nature. DFAs are particularly valued for their explanatory power due to their simplicity, minimal parameter usage, and the possibility of visualization, making them accessible and comprehensible to humans. On the other hand, the performance of transformers on formal languages remains somewhat enigmatic, as they generally underperform compared to state-based machine learning models like RNNs and LSTMs. This discrepancy has been documented in several research studies \cite{bhattamishra2020ability,Hahn_2020}, yet explanations rooted in the actual mechanisms have been lacking. In this context, we provide insights into how transformers operate on formal languages by analyzing the extracted DFAs, which have demonstrated high consistency in our extraction process, as detailed in Section~\ref{effectiveness}.

For regular languages that are fully learnable by transformers, extracting a DFA offers a straightforward explanation: the automaton is precisely correct, representing exactly what the transformer has learned, and providing the most accurate level of explanation possible. The scenario becomes more complex with languages that are not entirely learnable.

For the regular expressions in our training set, each corresponds to a simple DFA that can check a word's membership, typically involving fewer than 10 states, reflecting the simplicity of the grammar for human understanding. Previous research on extracting automata from state-based models like RNNs and LSTMs~\cite{weiss2020extracting} indicates that the simplicity of DFAs is generally preserved in the automata extracted from these models. However, this simplicity is not maintained with transformers, as evidenced by our experimental results.

For languages that transformers struggle to generalize, the extracted automata are significantly larger, with state counts around 30, as illustrated in Figure~\ref{fig:automaton-mod-3} and Figure~\ref{fig:automaton-tomita-3}. This disparity offers insights into why transformers underperform with formal languages. When the language's pattern becomes slightly complex, such as involving number counting, the attention mechanism fails to process the sequence token-by-token effectively. Moreover, transformers lack a concept of state similar to traditional models, relying instead on the QKV (Query, Key, Value) cache \cite{Attention}. In such cases, transformers do not genuinely generalize but tend to overfit more common long-term patterns by memorizing them in their QKV embeddings. This tendency leads them to memorize specific sequences rather than abstract the underlying, simpler patterns. The voluminous state counts in the extracted automata serve as evidence of this behavior, suggesting that the large, complex patterns produced are a byproduct of this memorization rather than genuine learning.

This extraction method not only sheds light on the specific challenges transformers face with formal languages but also supports the hypothesis that their poor performance may stem from an over-reliance on memorization of overall context which omits the important local patterns, as indicated by the complex and large automata derived from them.

\begin{center}
\begin{tcolorbox}[colback = white!100,width=\linewidth]
    {\textbf{Answer to RQ3:} Our method elucidates the behavior of encoder-only transformers on formal languages by extracting DFAs, which visually and structurally represent what the transformer has learned. This extraction provides clear insights, particularly highlighting the challenges transformers face with complex patterns and their tendency to overfit rather than generalize, evidenced by the significantly larger and more complex DFAs for languages that are difficult for transformers to learn.}
\end{tcolorbox}
\end{center}

\subsection{Ablation (RQ4)}
\begin{center}
\begin{tcolorbox}[colback = black!10,width=\linewidth]
    {Is it \textit{necessary} to introduce the representation-based technique into this extraction process?}
\end{tcolorbox}
\end{center}
One common concern in the pipeline is ensuring that the internal state-level consistency is maintained between models during the two-step extraction process, which is crucial for the extracted automaton to accurately reflect the patterns learned by the model. For the second extraction step, from DCSA to DFA, this issue is less problematic as the process involves abstracting the state space into discrete states, thereby maintaining consistency. However, it remains uncertain whether the DCSA has effectively generalized the pattern learned by the transformer. Evaluating the consistency rate \(C(T,D)\) could potentially assess this, though it is important to note that if the transformer fails to learn the language pattern, the DCSA cannot learn it precisely either, making perfect accuracy unattainable. Training the DCSA solely on the dataset labeled by the transformer may result in learning superficial connections rather than the precise transitions occurring within the transformer. To address this, we propose aligning the transformer's representation space with the DCSA's state space, treating the former as equivalent to its state space. By penalizing differences in these spaces during training, this alignment ensures consistency in their internal mechanisms, providing a robust justification for the effectiveness of leveraging representation alignment. This approach demonstrates dual benefits, which justifies our assertion: 1) it reduces the discrepancy between the representation space and the state space, and 2) it enhances the effectiveness of the extraction process.

For the first part of the justification, we define the average difference between representation space and its corresponding state space in DCSA within $L_p$-norm as the average difference of the corresponding $L_p$ distance of representation and state in the training dataset $X$, formally

\begin{equation}\begin{aligned}
\textbf{Diff}_p = \frac{1}{|X|}\sum_{x\in X} ||\text{Rep}(x) - \text{State}(x)||_{L^p}.
\end{aligned}\end{equation}

 We evaluate this difference for DCSA extracted from a transformer trained on Tomita 3 grammar. Results are shown in Table~\ref{tab:ablation_diff}. As indicated in the table, the inconsistency between the state space and the representation space is significantly reduced by the representation aligning mechanism for both \(L_1\) and \(L_2\) norms. This demonstrates that our method effectively maintains consistency in the inner state space concept.

To demonstrate that aligning representations practically impro\-ves extraction performance, we conducted an ablation study on some appropriately challenging classification tasks for the transformer. We measured the overall consistency \(C(T,A)\), with results displayed in Table~\ref{tab:ablation_effect}. According to the table, extraction performance is enhanced when the DCSA is trained with the representation aligning mechanism. In contrast, omitting this mechanism leads to diminished or even absent performance. This provides practical evidence supporting the effectiveness of our proposed representation alignment approach.

\begin{table}[]
    \centering
\begin{tabular}{c|c|c}
    \toprule
    Space Difference($\downarrow$) & w/ Rep Alignment & w/o Rep Alignment \\
    \midrule
    $\textbf{Diff}_1$ & \textbf{1.586} & 41.825 \\
    $\textbf{Diff}_2$ & \textbf{0.637} & 9.069\\
    \bottomrule
    
\end{tabular}
    \caption{Evaluation results of difference between the representation space of given transformer and corresponding state space with/without representation aligning when training DCSA on Tomita 3 dataset. Values in bold indicate better consistency.}
    \label{tab:ablation_diff}
\end{table}

\begin{table}[]
    \centering
\begin{tabular}{c|c|c}
    \toprule
    Grammar & w/ Rep Alignment & w/o Rep Alignment \\
    \midrule
    Tomita 3 & \textbf{0.9731} & 0.9260 \\
    Tomita 6 & \textbf{0.7603} & 0.5104\\
    $\mathcal{D}_2$ & \textbf{0.9899} & 0.9377   \\
    $\mathcal{D}_4$ & \textbf{0.9536} & 0.8965  \\
    \bottomrule
    
\end{tabular}
    \caption{Evaluation results of $C(T,A)$ on extracting DFA with/without representation aligning when training DCSA.}
    \label{tab:ablation_effect}
\end{table}

\begin{center}
\begin{tcolorbox}[colback = white!100,width=\linewidth]
    {\textbf{Answer to RQ4:} As shown in the ablation study, introducing the representation-based technique into the extraction process \textit{is} necessary as it significantly enhances the alignment and consistency between the model's internal representation space and state space, thereby improving the effectiveness of the extraction process.}
\end{tcolorbox}
\end{center}
\section{Related Work}
\label{chap:rw}

\textbf{Automata Extraction.}
The \textit{L}$^*$ algorithm, introduced by Angluin~\cite{angluin1987learning}, represents the pioneering method for learning automata from regular languages. However, since it relies solely on queries without utilizing any underlying models, it is more accurately described as a learning algorithm rather than an extraction algorithm. Giles et al.~\cite{giles1992learning} were the first to propose a specific DFA extraction algorithm, choosing to derive DFA from RNNs using a breadth-first search approach. This line of work was further advanced by Weiss et al.~\cite{weiss2020extracting}, who extended it to general state-based machine learning models such as LSTMs~\cite{sak2014long} and GRUs~\cite{chung2014empirical}. Later, Xu et al.~\cite{xu2021extracting} developed a method to expand this approach to all neural networks across various tasks using active learning. Nonetheless, these methodologies do not readily apply to transformers due to their fundamentally different architecture, which lacks a clear concept of state essential for traditional extraction methods.

Another line of research focuses on extracting weighted finite automata (WFA) from recurrent neural networks, as detailed in the foundational text by Droste et al.~\cite{droste2009handbook}. Okudono et al.~\cite{okudono2020weighted} pioneered a method to extract WFAs from RNNs using regression analyses on state spaces combined with the \textit{L}$^*$ algorithm. Zhang et al.~\cite{zhang2021decision} further advanced this area by introducing a technique to derive WFAs from RNNs through decision guidance. Wei et al.~\cite{wei2022extracting,wei2024weighted} expanded upon these methodologies to the domain of natural languages, employing empirical methods to fill gaps in the transition diagrams. However, these approaches are primarily tailored to address challenges in informal languages and do not readily adapt to the unique architecture of transformers. Additionally, WFAs are often heavily parameterized and maintain some degree of opacity, which limits their utility in explanatory contexts. Given the complexities associated with formal languages and regular expressions, this body of work does not provide an optimal solution for our specific challenges.

\textbf{Transformers for Formal Language.}
Recently, the significant advancements of transformers in natural language processing have sparked interest in their applicability to formal languages. However, the outcomes in this area are less promising. Hahn et al.\cite{Hahn_2020} highlighted substantial theoretical limitations in the computational capabilities of self-attention, demonstrating its inability to model periodic finite-state languages or hierarchical structures without increasing the number of layers or heads in proportion to input length. Further, Hao et al.\cite{10.1162/tacl_a_00490} conducted a refined analysis on various formal models of Transformer encoders, differing in their self-attention mechanisms, and outlined stricter theoretical constraints on the language complexity class that transformers can recognize. Bhattamishra et al.~\cite{bhattamishra2020ability} also investigated the capacity of transformers to model formal languages by training them to recognize various well-known formal languages, providing insights into the impact of positional encoding. While these studies are pivotal in identifying the challenges transformers face with formal languages and contribute theoretical perspectives, they do not offer practical methods to explain the behavior of transformers trained on specific formal languages or provide actionable insights into the practical issues transformers encounter with these languages.

\section{Conclusion}
In this paper, we have presented a pioneering, fully automated method for extracting deterministic finite automata from encoder-only transformers using representation-based abstraction and \textit{L}$^*$ algorithm, enhancing the interpretability of these complex neural network models as they process formal languages. Our method rigorously evaluates the effectiveness, adaptability, and explanatory power of the extracted automata, demonstrating its capability to reveal intricate patterns and operational dynamics hidden within transformer architectures and proving the effectiveness of the proposed representation alignment mechanism. This contribution is particularly valuable in the context of deep learning, where the black-box nature of models often obscures their underlying mechanisms. By bridging the gap between formal language processing and transformer understanding, our approach sets a foundation for future research in this area and opens up new avenues for comprehensively understanding machine learning models.

\section*{Acknowledgement}
This work was sponsored by the National Natural Science Foundation of China (Grant No. 62172019) and the Beijing Natural Science Foundation’s Undergraduate Initiating Research Program (Grant No. QY23041).
\newpage
\bibliographystyle{ACM-Reference-Format}
\bibliography{ref}


\end{document}